\documentclass[10pt,twocolumn,letterpaper]{article}

\usepackage{arxiv}
\usepackage{times}
\usepackage{epsfig}
\usepackage{graphicx}
\usepackage{amsmath}
\usepackage{amssymb}
\usepackage[breaklinks=true,bookmarks=false]{hyperref}
\usepackage{comment}
\usepackage{array}
\usepackage{booktabs}
\usepackage{xcolor}
\usepackage{lipsum}
\usepackage{multirow}
\usepackage{placeins}

\usepackage{soul}
\usepackage{amsmath}
\usepackage{amssymb}
\usepackage{microtype}
\usepackage{wrapfig}
\usepackage{fancyhdr}

\def\figurePath{figures_arxiv/}
\def\myfigure#1#2{\begin{figure}[htb]\centering\includegraphics*[width = \linewidth]{\figurePath#1}\centering\caption{#2}\label{fig:#1}\end{figure}}

\def\mycfigure#1#2{\begin{figure*}[t]\centering\includegraphics*[clip, width = \linewidth]{\figurePath#1}\centering\caption{#2}\label{fig:#1}\end{figure*}}
\newcommand{\mywrapfigure}[3]{%
\begin{wrapfigure}{r}{#2\columnwidth}%
  \begin{center}%
    \vspace{-0.5cm}%
    \includegraphics[width=#2\columnwidth]{\figurePath#1}%
    \caption{#3}%
    \vspace{-0.5cm}%
    \label{fig:#1}%
  \end{center}%
\end{wrapfigure}%
\leavevmode%
}

\renewcommand{\eg}{e.\,g., }
\renewcommand{\ie}{i.\,e., }
\renewcommand{\etal}{et~al.\ }

\newcommand{\refSec}[1]{Sec.~\ref{sec:#1}}
\newcommand{\refFig}[1]{Fig.~\ref{fig:#1}}
\newcommand{\refEq}[1]{Eq.~\ref{eq:#1}}
\newcommand{\refTbl}[1]{Tbl.~\ref{tbl:#1}}

\soulregister\ref7
\soulregister\cite7
\soulregister\refFig7
\soulregister\cite7
\soulregister\ref7
\soulregister\pageref7
\soulregister\shortcite7
\soulregister\eg0
\soulregister\ie0
\soulregister\etal0

\DeclareGraphicsExtensions{.png,.jpg,.pdf,.ai,.psd}
\newcommand{\mysection}[2]{\section{#1}\label{sec:#2}}
\newcommand{\mysubsection}[2]{\subsection{#1}\label{sec:#2}}

\usepackage{pifont}
\newcommand{\cmark}{\checkmark}%
\newcommand{\xmark}{\scalebox{0.85}{\ding{53}}}%
\usepackage{adjustbox}
\newcolumntype{R}[2]{%
    >{\adjustbox{angle=#1}\bgroup}%
    l%
    <{\egroup}%
}
\newcommand*\rot[2]{\multicolumn{1}{R{#1}{#2}}}%

\newcommand{\exemplar}{\mathbf y}
\newcommand{\decoder}{f}
\newcommand{\encoder}{g}
\newcommand{\translator}{h}
\newcommand{\sampler}{s}
\newcommand{\tunableParameters}{\theta}
\newcommand{\tunableEncoderParameters}
    {\tunableParameters_{\mathrm g}}
\newcommand{\tunableTranslatorParameters}
    {\tunableParameters_{\mathrm h}}
\newcommand{\tunableDecoderParameters}
    {\tunableParameters_{\mathrm d}}
\newcommand{\textureCode}{\mathbf z}
\newcommand{\decoderParams}{\mathbf e}
\newcommand{\position}{{\mathbf x}}
\newcommand{\n}{\mathbf{n}}
\newcommand{\noise}{\mathtt{noise}}

\newcommand{\weight}{w}
\newcommand{\numberOfOctaves}{m}
\newcommand{\mlp}{\mathtt{mlp}}
\newcommand{\cnn}{\mathtt{cnn}}
\newcommand{\transform}{\mathsf T}
\newcommand{\seed}{\xi}

\definecolor{perlinColor}{HTML}{557EBF}
\definecolor{perlinTColor}{HTML}{EA4535}
\definecolor{cnnColor}{HTML}{F9BC15}
\definecolor{cnnDColor}{HTML}{36A852}
\definecolor{mlpColor}{HTML}{F36E21}
\definecolor{oursPColor}{HTML}{48BEC6}
\definecolor{oursNoTColor}{HTML}{A8C1E5}
\definecolor{oursColor}{HTML}{000000}

\begin{document}

\title{Learning a Neural 3D Texture Space from 2D Exemplars}

\author{Philipp Henzler$^{1}$\\
{\tt\small p.henzler@cs.ucl.ac.uk}
\\
\and
Niloy J. Mitra$^{1,2}$\\
{\tt\small n.mitra@cs.ucl.ac.uk}\vspace{.3cm}\\
\and
Tobias Ritschel$^{1}$\\
{\tt\small t.ritschel@ucl.ac.uk}
\\
\and
$^{1}$University College London \and
$^{2}$Adobe Research
}

\maketitle
\thispagestyle{empty}
\begin{abstract}
We propose a generative model of 2D and 3D natural textures with diversity, visual fidelity and at high computational efficiency.
This is enabled by a family of methods that extend ideas from classic stochastic procedural texturing (Perlin noise) to learned, deep, non-linearities.
The key idea is a hard-coded, tunable and differentiable step that feeds multiple transformed random 2D or 3D fields into an MLP that can be sampled over infinite domains.
Our model encodes all exemplars from a diverse set of textures without a need to be re-trained for each exemplar.
Applications include texture interpolation, and learning 3D textures from 2D exemplars. 
Project website: \url{https://geometry.cs.ucl.ac.uk/projects/2020/neuraltexture}.
\end{abstract}

\mysection{Introduction}{Introduction}
Textures are stochastic variations of attributes over 2D or 3D space with applications in both image understanding and synthesis.
This paper suggests a generative model of natural textures. Previous texture models either capture a single exemplar (\eg wood) alone or address non-stochastic (stationary) variation of appearance across space: Which location on a chair should have a wood color? Which should be cloth? Which metal?
Our work combines these two complementary views.

\myfigure{Teaser}{Our approach allows casually-captured 2D textures (blue) to be mapped to latent texture codes and support interpolation (blue-to-red), projection, or synthesis of volumetric textures.}

\paragraph{Requirements}
We design the family of methods with several requirements in mind: completeness, generativeness, compactness, interpolation, infinite domains, diversity, infinite zoom, and high speed.

A space of textures is \emph{complete}, if every natural texture has a compact code $\textureCode$ in that embedding.
To be \emph{generative}, every texture code should map to a useful texture.
This is important for intuitive design where a user manipulates the texture code and expects the outcome to be a texture.
\emph{Compactness} is achieved if codes are low-dimensional.
We also demand the method to provide \emph{interpolation}: texture generated at coordinates between $\textureCode_1$ and $\textureCode_2$ should also be valid.
This is important for design or when storing texture codes into a (low-resolution) 2D image, 3D volume or at mesh vertices with the desire to interpolate.
The first four points are typical for generative modelling; achieving them jointly while meeting more texture-specific requirements (stochasticity, efficiency) is our key contribution.

First, we want to support \emph{infinite domains}: Holding the texture code $\decoderParams$ fixed, we want to be able to query this texture so that a patch around any position $\position$ has the statistics of the exemplar.
This is important for querying textures in graphics applications for extended virtual worlds, \ie grass on a football field where it extends the size of the texture.

Second, for visual fidelity, the statistics under which textures are \emph{similar} to the exemplar.
The Gram matrix of VGG activations is one established metric for this similarity \cite{gatys2015texture}.

Third, \emph{infinite zoom} means each texture should have variations on a wide range of scales and not be limited to any fixed resolution that can be held in memory.
This is required to zoom into details of geometry and appreciate the fine variation such as wood grains, etc.
In practice, we are limited by the frequency content of the exemplars we train on, but the method should not impose any limitations across scales.

Fourth and finally, our aim is \emph{computational efficiency}: the texture needs to be queryable without requiring prohibitive amounts of memory or time, in any dimension.
Ideally, it would be constant in both and parallel.
This rules out simple convolutional neural networks, that do not scale favorable in memory consumption to 3D.

\mysection{Previous Work}{PreviousWork}
Capturing the variations of nature using stochastic on many scales has a long history \cite{mandelbrot1983fractal}.
Making noise useful for graphics and vision is due to Perlin's 1995 work \cite{perlin1985noise}.
Here, textures are generated by computing noise at different frequencies and mixing it with linear weights.
A key benefit is that this noise can be evaluated in 2D as well as in 3D making it popular for many graphics applications.

Computer vision typically had looked into generating textures from exemplars, such as by non-parametric sampling \cite{efros1999texture}, vector quantization \cite{wei2000fast}, optimization \cite{kwatra2005texture} or nearest-neighbor field synthesis (PatchMatch \cite{barnes2009patchmatch}) with applications in in-painting and also (3D) graphics.
Typically, achieving spatial and temporal coherence as well as scalability to fine spatial details remains a challenge.
Such classic methods cater to the requirements of human texture perception as stated by Julesz \cite{julesz1965texture}: a texture is an image full of features that in some representation have the same statistics.

The next level of quality was achieved when representations became learned, such as the internal activations of the VGG network \cite{simonyan2014very}.
Neural style transfer \cite{gatys2015texture} looked into the statistics of those features, in particular, their Gram matrices.
By optimizing over pixel values, these approaches could produce images with the desired texture properties.
If these properties are conditioned on existing image structures, the process is referred to as style transfer.
VGG was also used for optimization-based multi-scale texture synthesis \cite{sendik2017deep}. Such methods require optimizations for each individual exemplar.

Ulyanov et al.~\cite{ulyanov2016texture} and Johnson et al.~\cite{johnson2016perceptual} have proposed networks that directly produce the texture without optimization.
While now a network generated the texture, it was still limited to one exemplar, and no diversity was demonstrated.
However, noise at different resolutions
\cite{perlin1985noise} is input to these methods, also an inspiration to our work.
Follow up work \cite{ulyanov2017improved} has addressed exactly this difficulty by introducing an explicit diversity term \ie asking all results in a batch to be different.
Unfortunately, this frequently introduces mid-frequency oscillations of brightness that appear admissible to VGG instead of producing true diversity.
In our work, we achieve diversity, by restricting the networks input to stochastic values only, \ie diversity-by-construction

A certain confusion can be noted around the term ``texture''.
In the human vision \cite{julesz1965texture} and computer vision literature \cite{efros1999texture,hertzmann2001image}, it exclusively refers to stochastic variation.
In computer graphics, \eg OpenGL, ``texture'' can model both stochastic and non-stochastic variation of color.
For example, Visual Object Networks \cite{zhu2018visual} generate a voxel representation of shape and diffuse albedo and refer to the localized color appearance, \eg wheels of a car are dark, the rim are silver, etc., as ``texture''.
Similar, Oechsle et al.~\cite{oechsle2019texturefields} and Saito \etal~\cite{saito2019pifu} use an implicit function to model this variation of appearance in details beyond voxel resolution.
Our comparison will show, how methods tackling space of non-stochastic texture variation \cite{oechsle2019texturefields,zhu2018visual}, unfortunately are not suitable to model stochastic appearance.
Our work is progress towards learning spaces of stochastic and non-stochastic textures.

Some work has used adversarial training to capture the essence of textures \cite{shaham2019singan, bergmann2017learning}, including the non-stationary case \cite{zhou2018non} or even inside a single image \cite{shaham2019singan}.
In particular StyleGAN \cite{karras2019style} generates images with details by transforming noise in adversarial training.
We avoid the challenges of adversarial training but train a NN to match VGG statistics.

Aittala et al.~\cite{aittala2016reflectance} have extended Gatsy et al.'s 2015 \cite{gatys2015texture} approach to not only generate color, but also ensembles of 2D BRDF model parameter maps from single 2D exemplars.
Our approach is compatible with this approach, for example to generate 3D bump, specular, etc. maps, but from 2D input.

At any rate, none of the texture works in graphics or vision \cite{perlin1985noise,gatys2015texture,ulyanov2016texture,efros1999texture,barnes2009patchmatch,xian2018texturegan,yu2019texture} generate a space of textures, such as we suggest here, but all work on a single texture while the ones that work on a space of exemplars \cite{zhu2018visual,oechsle2019texturefields} do not create stochastic textures.
Our work closes this gap, by creating a space of stochastic textures.

The graphics community, however, has looked into generating spaces of textures \cite{matusik2005texture}, which we here revisit from a deep learning perspective.
Their method deforms all pairs of exemplars to each other and constructs a graph with edges that are valid for interpolation when there is evidence that the warping succeeded.
To blend between them, histogram adjustments are made.
Consequently, interpolation between exemplars is not a straight path from one another, but a traversal along valid observations. Similarly, our method could also construct valid paths in the latent space interpolation.

Finally, all these methods require to learn the texture in the same space it will be used, while our approach can operate in any dimension and across dimensions, including the important case of generating procedural 3D solid textures from 2D observations \cite{kopf2007solid} or slices \cite{pietroni2007texturing} only.

\paragraph{Summary}
\begin{table}[h]
    \centering
    \caption{%
    Comparison of texture synthesis methods.
    Please see text for refined definition of the rows and columns.
    }
    \label{tbl:PreviousWork}
    \setlength{\tabcolsep}{2.0pt}
    \begin{tabular}{llllllllll}
        &
        Method&
        &
        \rot{90}{2em}{\scriptsize Diverse}&
        \rot{90}{2em}{\scriptsize Details}&
        \rot{90}{2em}{\scriptsize Speed}&
        \rot{90}{2em}{\scriptsize 3D}&
        \rot{90}{2em}{\scriptsize Quality}&
        \rot{90}{2em}{\scriptsize Space}&
        \rot{90}{2em}{\scriptsize2D-to-3D}
        \\
        \toprule
        \textcolor{perlinColor}{\textbullet}&
        Perlin&
        \texttt{perlin}&
        \cmark&
        \cmark&
        \cmark&
        \cmark&
        \xmark&
        \xmark&
        \xmark\\
        \textcolor{perlinTColor}{\textbullet}&
        Perlin + transform&
        \texttt{perlinT}&
         \cmark&
         \cmark&
         \cmark&
         \cmark&
         \xmark&
         \xmark&
         \xmark\\
        \textcolor{cnnColor}{\textbullet}&
        CNN&
        \texttt{cnn}&
         \xmark&
         \xmark&
         \xmark&
         \xmark&
         \cmark&
         \xmark&
         \xmark\\
        \textcolor{cnnDColor}{\textbullet}&
        CNN + diversity&
        \texttt{cnnD}&
         \cmark&
         \xmark&
         \xmark&
         \xmark&
         \xmark&
         \xmark&
         \xmark\\
        \textcolor{mlpColor}{\textbullet}&
        MLP&
        \texttt{mlp}&
         \xmark&
         \xmark&
         \cmark&
         \cmark&
         \xmark&
         \xmark&
         \cmark\\
        \textcolor{oursPColor}{\textbullet}&
        Ours + position&
        \texttt{oursP}&
         \xmark&
         \cmark&
         \cmark&
         \cmark&
         \xmark&
         \cmark&
         \cmark\\
        \textcolor{oursNoTColor}{\textbullet}&
        Ours - transform&
        \texttt{oursNoT}&
         \xmark&
         \xmark&
         \cmark&
         \cmark&
         \cmark&
         \cmark&
         \cmark\\
        \textcolor{oursColor}{\textbullet}&
        Ours&
        \texttt{ours}&
         \cmark&
         \cmark&
         \cmark&
         \cmark&
         \cmark&
         \cmark&
         \cmark\\         
         \bottomrule
    \end{tabular}
\end{table}
The state of the art is depicted in \refTbl{PreviousWork}. Rows list different methods while columns address different aspects of each method.
A method is ``Diverse'' if more than a single exemplar can be produced.
MLP \cite{oechsle2019texturefields} is not diverse as the absolute position allows overfitting.
We denote a method to have ``Detail'' if it can produce features on all scales.
CNN does not have details, as, in particular in 3D, it needs to represent the entire domain in memory, while MLPs and ours are point operations.
``Speed'' refers to computational efficiency.
Due to high bandwidth and lacking data parallelism, a CNN, in particular in 3D, is less efficient than ours.
This prevents application to ``3D''.
``Quality'' refers to visual fidelity, a subjective property.
CNN, MLP and ours achieve this, but Perlin is too simple a model.
CNN with diversity \cite{ulyanov2017improved} have decent quality, but a step back from \cite{ulyanov2016texture}.
Our approach creates a ``Space'' of a class of textures, while all others only work with single exemplars.
Finally, our approach allows to learn from a single 2D observation \ie 2D-to-3D.
MLP \cite{oechsle2019texturefields} also learn from 2D images, but have multiple images of one exemplar, and pixels are labeled with depth.

\mysection{Overview}{Overview}
Our approach has two steps.
The first embeds the exemplar into a latent space using an \emph{encoder}.
The second provides \emph{sampling} at any  position by reading noise fields at that position and combining them using a learned mapping to match the exemplar statistics.
We now detail both steps.

\myfigure{Overview}{Overview of our approach as explained in \refSec{Overview}.}%
\vspace*{-.1in}

\paragraph{Encoder}
The encoder $\encoder$ maps a 2D texture exemplar image $\exemplar$  to a latent texture code $\textureCode=\encoder(\exemplar)$.
We use a convolutional neural network to encode the high number of exemplar pixels into a compact latent texture code $\textureCode$.

\paragraph{Sampler}
Sampling $\sampler(\position|\textureCode)$ of a texture with code $\textureCode$ at individual 2D or 3D positions $\position$ has two steps: a \emph{translator} and a \emph{decoder}, which are both described next.

\paragraph{Decoder}
Our key idea is to prevent the decoder $\decoder(\n|\decoderParams)$ to access the position $\position$ and to use a vector of noise values $\n$ instead.
Each $
n_i = 
\noise(
\transform_i
2^{i-1}
\position
|
\seed_i)
$ is read at different linear transformations $\transform_i2^{i-1}\position$ of that position $\position$ from random fields with different seeds $\seed_i$.
The random field $\noise(\position | \seed_i)$ is implemented as an infinite, single-channel 2D or 3D function that has the same random value for all continuous coordinates $\position$ in each integer lattice cell for one seed $\seed_i$.
The factors of $2^{i-1}$ initialize the decoder to behave similar to Perlins's octaves for identity $\transform_i$.
Applying $\transform_i 2^{i-1}$ to $\position$ is similar to Spatial Transformer Networks \cite{jaderberg2015spatial}.
 (\refFig{sample_noise}). 
\myfigure{sample_noise}{Noise field for different octaves and transformations $\transform$.}

These noise values are combined with the extended texture code $\decoderParams$ in a learned way.
It is the task of the translator, explained next, to control, given the exemplar, how noise is transformed and to generate an extended texture code.

\paragraph{Translator}
\mywrapfigure{translator}{0.50}{Translator.}
The translator $\translator(\textureCode)=\{ \decoderParams, \transform \}$
maps the texture code $\textureCode$ to a tuple of parameters required by the decoder: the vector of transformation matrices $\transform$ and an extended texture code vector $\decoderParams$.
The matrices $\transform$ are used to transform the coordinates before reading the noise as explained before.
The extended texture parameter code $\decoderParams$ is less compact than the texture code $\textureCode$, but allows the sampler to execute more effectively, \ie do not repeat computations required for different $\position$ as they are redundant for the same $z$. 

See \refFig{translator} where for example two $2 \times 2 $ transformation matrices with 8 DOF are parameterized by three parameters.
 
\paragraph{Training}
For training, the encoder is fed with a random $128 \times 128$ patch $P_\mathrm e$ of a random exemplar $\exemplar$, followed by the sampler evaluating a regular grid of $128 \times 128$ points $\position$ in random 2D slices of the target domain to produce a ``slice'' image $P_\mathrm s$ (\refFig{Slicing}).
The seed $\seed$ is held constant per train step, as one lattice cell will map to multiple pixels, and the decoder $\decoder$ relies on these being consistent. 
During inference changing the seed $\seed$ and keeping the texture code $\decoderParams$ will yield diverse textures.
\myfigure{Slicing}{
 Sliced loss for learning 3D procedural textures from 2D exemplars:
 Our method, as it is non-convolutional, can sample the 3D texture (a) at arbitrary 3D positions.
 This enables to also sample arbitrary 2D slices (b).
 For learning, this allows to simply slice 3D space along the three major axes (red, yellow, blue) and ask each slice to have the same VGG statistics as the exemplar (c).
}

The loss is the $\mathcal L_2$ distance of Gram matrix of VGG feature activations \cite{gatys2015texture,johnson2016perceptual,ulyanov2017improved,ulyanov2016texture,aittala2016reflectance} of the patches $P_\mathrm e$ and $P_\mathrm s$.

If the source and target domain are the same (synthesizing 2D textures from 2D exemplars) the slicing operation is the identity.
However, it also allows for the important condition in which the target domain has more dimensions than the source domain, such as learning 3D from 2D exemplars.
 
\paragraph{Spaces-of}
Our method can be used to either fit a \emph{single} exemplar or an entire \emph{space} of textures.
In the single mode, we directly optimize for the trainable parameters $
\tunableParameters=
\{
\tunableDecoderParameters
\}$ of the decoder.
When learning the entire space of textures, the full cascade of encoder $\encoder$, translator $\translator$ and sampler $\sampler$ parameters are trained, \ie $
\tunableParameters=
\{
\tunableEncoderParameters,
\tunableTranslatorParameters,
\tunableDecoderParameters
\}$ jointly.

\mysection{Learning stochastic space coloring}{OurMethod}
Here we will introduce different implementations of samplers $\sampler \colon \mathbb{R}^n \to \mathbb{R}^3$ which ``color'' 2D or 3D space at position $\position$. 
We discuss pros and cons with respect to the requirements from the introduction, ultimately leading to our approach.

\paragraph{Perlin} noise is a simple and effective method to generate natural textures in 2D or 3D \cite{perlin1985noise}, defined as
\begin{align}
\begin{split}
\label{eq:Perlin}
\sampler(\position|\textureCode) &= 
\sum_{i=1}^{\numberOfOctaves}
\noise(2^{i-1}\position, \seed_i) \otimes \weight_{i},
\end{split}
\end{align}
where $\translator(\textureCode) = \{\weight_1, \weight_2, \ldots \} $ are the RGB weights for $\numberOfOctaves$ different noise functions $\noise_i$ which return bilinearly-sampled RGB values from an integer grid. $\otimes$ is channel-wise multiplication.
Here, $\decoderParams$ is a list of all linear per-layer RGB weights \eg an 8$\times$3 vector for the $\numberOfOctaves=8$ octaves we use.
This is a simple latent code, but we will see increasingly complex ones later.
Also our encoder $\encoder$ is designed such that it can cater to all decoders, even Perlin noise \ie we can also create a space of textures with a Perlin noise back-end.

Coordinates $\position$ are multiplied by factors of two (octaves), so with increasing $i$, increasingly smooth noises are combined.
This is motivated well in the spectra of natural signals \cite{mandelbrot1983fractal,perlin1985noise}, but also limiting.
Perlin's linear scaling allows the noise to have different colors, yet no linear operation can reshape a distribution to match a target.
Our work seeks to overcome these two limitations, but tries to retain the desirable properties of Perlin noise: simplicity and computational efficiency as well as generalization to 3D. 

\paragraph{Transformed Perlin} relaxes the scaling by powers of two
\begin{align}
\begin{split}
\label{eq:PerlinTransf}
\sampler(\position|\textureCode) &= 
\sum_{i=1}^{\numberOfOctaves}
\noise(\transform_i2^{i-1}\position, \seed_i) \otimes \weight_{i}
\end{split}
\end{align}
by allowing each noise $i$ to be independently scaled by its own transformation matrix $\transform_i$ since $\translator(\textureCode) = \{\weight_1, \transform_1, \weight_2, \transform_2, \ldots \} $.
Please note, that the choice of noise frequency is now achieved by scaling the coordinates reading the noise.
This allows to make use of anisotropic scaling for elongated structures, different orientations or multiple random inputs at the same scale.

\paragraph{CNN} utilizes the same encoder $\encoder$ as our approach to generate a texture code that is fed in combination with noise to a convolutional decoder similar to \cite{ulyanov2017improved}.

\begin{align}
\begin{split}
\label{eq:CNN}
\sampler(\position|\textureCode) &= 
\cnn(\position|\decoderParams, \noise(\seed))
\end{split}
\end{align}
The CNN is conditioned on $\decoderParams$ without additional translation. Their visual quality is stunning, CNNs are powerful and the loss is able to capture perceptually important texture features, hence CNNs are a target to chase for us in 2D in terms of quality.
However, there are two main limitations of this approach we seek to lift: efficiency and diversity.

CNNs do not scale well to 3D in high resolutions.
To compute intermediate features at $\position$, they need to have access to neighbors.
While this is effective and output-sensitive in 2D, it is not in 3D: we need results for 2D surfaces embedded in 3D, and do so in spatial high resolution (say 1024$\times$1024), but this requires CNNs to compute a full 3D volume with the same order of pixels.
While in 2D partial outputs can be achieved with sliding windows, it is less clear how to slide a window in 3D, such that it covers all points required to cover all 3D points that are part of the visible surface.

The second issue is diversity: CNNs are great for producing a re-synthesis of the input exemplar, but it has not been demonstrated that changing the seed $\seed$ will lead to variation in the output in most classic works \cite{ulyanov2016texture,johnson2016perceptual} and in classic style transfer \cite{gatys2015texture} diversity is eventually introduced due to the randomness in SGD.
Recent work by Ulyanov and colleagues \cite{ulyanov2017improved} explicitly incentivizes diversity in the loss.
The main idea is to increase the pixel variance inside all exemplars produced in one batch.
Regrettably, this often is achieved by merely shifting the same one exemplar slightly spatially or introducing random brightness fluctuations.

\paragraph{MLP} maps a 3D coordinate to appearance:
\begin{align}
\begin{split}
\label{eq:MLP}
\sampler(\position|\textureCode) &= 
\mlp(\position|\decoderParams)
\end{split}
\end{align}
where $\translator(\textureCode)=\decoderParams$.
Texture-fields \cite{oechsle2019texturefields} have used this approach to produce what they call ``texture'', detailed and high-quality appearance decoration of 3D surfaces, but what was probably not intended is to produce diversity or any stochastic results. At least, there is no parameter that introduces any randomness, so all results are identical.
We took inspiration in their work, as it makes use of 3D point operations, that do not require accessing any neighbors and no intermediate storage for features in any dimensions, including 3D.
It hence reduces bandwidth compared to CNN, is perfectly data-parallel and scalable.
The only aspect missing to make it our colorization operator, required to create a space and evolve from 2D exemplars to 3D textures, is stochasticity.

\paragraph{Ours} combines the noise from transformed Perlin for stochasticity, the losses used in style and texture synthesis CNNs for quality as well as the point operations in MLPs for efficiency as follows:

\begin{alignat}{4}
\label{eq:Ours}
\sampler(\position|\textureCode)
= 
\decoder(
&\noise(\transform_1
&&2^0
&&\position, \seed_1
&&),
\ldots,\nonumber\\
&\noise(\transform_{\numberOfOctaves}
&&2^{m-1}
&&\position, \seed_{\numberOfOctaves}
&&)
|
\decoderParams)
\end{alignat}
Different from MLPs that take the coordinate $\position$ as input, position itself is hidden.
Instead of position, we take multiple copies of spatially smooth noise $\noise(\position)$ as input, with explicit control of how the noise is aligned in space expressed by the transformations $\transform$.
Hence, the MLP requires to map the entire distribution of noise values such that it suits the loss, resulting in build-in diversity. We chose number of octaves $\numberOfOctaves$ to be 8, \ie the transformation matrices $\transform_1,\ldots, \transform_{\numberOfOctaves}$ require $8 \times 4=32$ values in 2D. The texture code size $\decoderParams$ is 64 and the compact code $\textureCode$ is 8.
The decoder $\decoder$ consists of four stacked linear layers, with 128 units each followed by ReLUs. The last layer is 3-valued RGB.

\paragraph{Non-stochastic ablation} seeks to investigate what happens if we do not limit our approach to random variables, but also provide access to deterministic information $\position$:
\begin{alignat}{3}
\label{eq:OursNonStoch}
\sampler(\position|\textureCode)
= 
\decoder(
\position,
&\noise(
2^{0}
&&\position,
\seed_1
&&),
\ldots,\nonumber\\
&\noise(
2^{\numberOfOctaves-1}
&&\position,
\seed_{\numberOfOctaves}
&&)
|
\decoderParams)
\end{alignat}
is the same as MLP, but with access to noise.
We will see that this effectively removes diversity.

\paragraph{Non-transformed ablation} evaluates, if our method were to read only from multi-scale noise without control over how it is transformed.
Its definition
\begin{alignat}{3}
\label{eq:OursNonTransf}
\sampler(\position|\textureCode) = 
\decoder(
&\noise(2^{0}
&&\position, \seed_1
&&),
\ldots,\nonumber\\
&\noise(2^{\numberOfOctaves-1}
&&\position, \seed_{\numberOfOctaves}
&&)
|
\decoderParams)
\end{alignat}

\mycfigure{Quantitative}{%
Quantitative evaluation.
Each plot shows the histogram of a quantity (from top to bottom: success, style error and diversity) for different data sets (from left to right: all space together, \textsc{Wood}, \textsc{Marble}, \textsc{Grass}).
For a discussion, see the last paragraph in \refSec{Quantitative}.
}

\mysection{Evaluation}{Evaluation}
Our evaluation covers qualitative (\refSec{Quantitative}) and quantitative (\refSec{Qualitative}) aspects as well as a user study (\refSec{UserStudy}).

\mysubsection{Protocol}{Protocol}
We suggest a data set that for which we explore the relation of different methods, according to different metrics to quantify texture similarity and diversity.

\paragraph{Data set}
Our data set contains four classes 
(\textsc{Wood},
\textsc{Marble},
\textsc{Grass} and
\textsc{Rust})
of 2D textures, acquired from internet image sources. Each class contains 100 images.

\paragraph{Methods}
We compare eight different methods that are competitors, ablations and ours.

As five \emph{competitors} we study variants of Perlin noise, CNNs and MLPs.
\texttt{perlin} implements Perlin noise (\refEq{Perlin}, \cite{perlin1985noise}) and \texttt{perlinT} our variant extending it by a linear transformation (\refEq{PerlinTransf}).
Next, \texttt{cnn} is a classic TextureNet \cite{ulyanov2016texture} and \texttt{cnnD} the extension to incentivise diversity (\cite{ulyanov2017improved}, \refEq{CNN}).
\texttt{mlp} uses an MLP following \refEq{MLP}.

We study three \emph{ablations}.
First, we compare to \texttt{oursP} that is our method, but with the absolute position as input and no transform.
Second, \texttt{oursNoT} omits the absolute position as input and transformation but still uses Perlin's octaves (\refEq{OursNonTransf}).
The final method is \texttt{ours} method (\refEq{Ours}).

\paragraph{Metrics}
We evaluate methods in respect to three metrics: similarity and diversity and a joint measure, success.

\emph{Similarity} is high, if the result produced has the same statistics as the exemplar in terms of L2 differences of VGG Gram matrices.
This is identical to the loss used.
Similarity is measured on a single exemplar.

\emph{Diversity} is not part of the loss, but can be measured on a set of exemplars produced by a method.
We measure diversity by looking at the VGG differences between all pairs of results in a set produced for a different random seed.
Note, that this does not utilize any reference.
Diversity is maximized by generating random VGG responses, yet without similarity.

\emph{Success} of the entire method is measured as the product of diversity and the maximum style error minus the style error.
We apply this metric, as it combines similarity and diversity that are conflicting goals we jointly want to maximize.

\emph{Memory and speed} are measured at a resolution of 128 pixels/voxels on an Nvidia Titan Xp.

\mysubsection{Quantitative results}{Quantitative}

\begin{table}[ht]
    \centering
    \footnotesize
    \caption{%
    Efficiency in terms of compute time and memory usage in 2D and 3D \textbf{(columns)} for different methods \textbf{(rows)}.
    }
    \label{tbl:ComputationalEfficeincy}
    \begin{tabular}{rrrrr}
        \multicolumn{1}{c}{\multirow{2}{*}{Method}}&
        \multicolumn{2}{c}{Time}&
        \multicolumn{2}{c}{Memory}
        \\
        \cmidrule(lr){2-3}
        \cmidrule(lr){4-5}
        &
        \multicolumn{1}{c}{2D}&
        \multicolumn{1}{c}{3D}&
        \multicolumn{1}{c}{2D}&
        \multicolumn{1}{c}{3D}\\
        \toprule
\texttt{perlin} \textcolor{perlinColor}{\textbullet}
& 0.18\,ms & 0.18\,ms & 65\,k & 16\,M \\
\texttt{perlinT} \textcolor{perlinTColor}{\textbullet}&
0.25\,ms & 0.25\,ms & 65\,k & 16\,M \\
\texttt{cnn} \textcolor{cnnColor}{\textbullet}&
1.45\,ms & 551.59\,ms & 8,000\,k & 646\,M \\
\texttt{cnnD} \textcolor{cnnDColor}{\textbullet}&
1.45\,ms & 551.59\,ms & 8,000\,k & 646\,M \\
\texttt{mlp} \textcolor{mlpColor}{\textbullet}& 
1.43\,ms & 1.43\,ms & 65\,k & 16\,M \\
\midrule
\texttt{oursP} \textcolor{oursPColor}{\textbullet}&
1.44\,ms & 1.44\,ms & 65\,k & 16\,M \\
\texttt{oursNoT} \textcolor{oursNoTColor}{\textbullet}&
1.24\,ms & 1.24\,ms & 65\,k & 16\,M \\
\texttt{ours} \textcolor{oursColor}{\textbullet}&
1.55\,ms & 1.50\,ms & 65\,k & 16\,M \\   
         \bottomrule
    \end{tabular}
    \includegraphics[width = \linewidth]{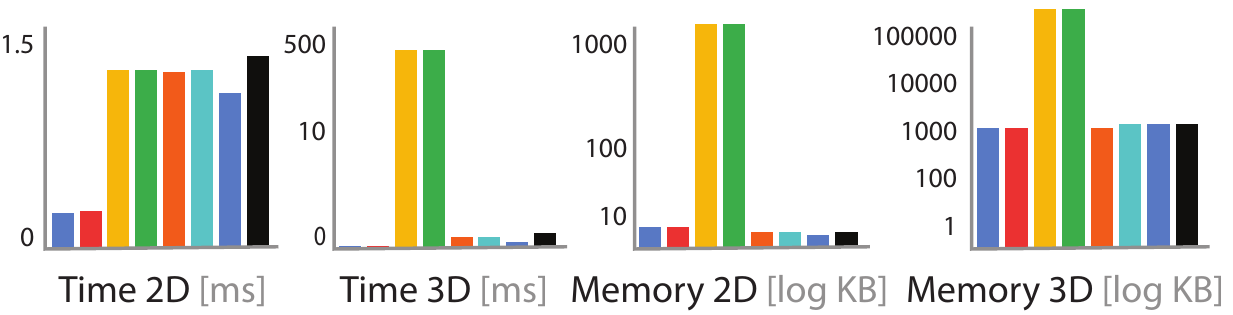}%
\end{table}

\paragraph{Efficiency}
We first look at computational efficiency in \refTbl{ComputationalEfficeincy}.
We see that our method shares the speed and memory efficiency with Perlin noise and MLPs / Texture Fields \cite{oechsle2019texturefields}.
Using a CNN \cite{ulyanov2016texture,ulyanov2017improved} to generate 3D textures as volumes is not practical in terms of memory, even at a modest resolution. Ours scales linear with pixel resolution as an MLP is a point-estimate in any dimension that does not require any memory other than its output.
A CNN has to store the internal activations of all layers in memory for information exchange between neighbors.

\begin{table}[t]
    \centering
    \setlength{\tabcolsep}{1.5pt}    
    \footnotesize
    \caption{%
    Similarity and diversity for methods on different textures.
    }
    \label{tbl:Quantitative}
    \begin{tabular}{r rrr rrr rrr rrr}
        \multicolumn{1}{c}{\multirow{2}{*}{Method}}&
        \multicolumn{3}{c}{\textsc{All}}&
        \multicolumn{3}{c}{\textsc{Wood}}&
        \multicolumn{3}{c}{\textsc{Grass}}&
        \multicolumn{3}{c}{\textsc{Marble}}
        \\
        \cmidrule(lr){2-4}
        \cmidrule(lr){5-7}
        \cmidrule(lr){8-10}
        \cmidrule(lr){11-13}
        &
        \multicolumn{1}{c}{Sim}&
        \multicolumn{1}{c}{Div}&
        \multicolumn{1}{c}{Suc}&
        \multicolumn{1}{c}{Sim}&
        \multicolumn{1}{c}{Div}&
        \multicolumn{1}{c}{Suc}&
        \multicolumn{1}{c}{Sim}&
        \multicolumn{1}{c}{Div}&
        \multicolumn{1}{c}{Suc}&
        \multicolumn{1}{c}{Sim}&
        \multicolumn{1}{c}{Div}&
        \multicolumn{1}{c}{Suc}
        \\
        \toprule
        \texttt{\tiny perlin} \textcolor{perlinColor}{\textbullet}&
        20.6&
        48.0&
        7.0&
        23.8&
        37.9&
        4.9&
        24.6 & 72.8 & 18.1&
        13.3 & 31.8 & 7.84
        \\
        \texttt{\tiny perlinT}
        \textcolor{perlinTColor}{\textbullet}&
        19.6&
        48.2&
        7.2&
        18.4&
        39.6&
        5.02&
        25.9 & 65.6 & 13.8&
        14.2 & 38.4 & 8.03
        \\
        \texttt{\tiny cnn}
        \textcolor{cnnColor}{\textbullet}&
        5.4&
        0.5&
        7.5&
        13.4&
        0.5&
        0.07&
        \textbf{1.9}&
        0.5&
        0.14&
        \textbf{1.1}&
        0.3&
        0.08
        \\
        \texttt{\tiny cnnD}
        \textcolor{cnnDColor}{\textbullet}&
        \textbf{3.9}&
        48.2&
        7.75&
        \textbf{3.9}&
        35.2&
        5.19&
        4.8&
        59.2&
        20.9&
        3.6&
        48.8&
        8.5
        \\
        \texttt{\tiny mlp}
        \textcolor{mlpColor}{\textbullet}&
        14.1&
        0.0&
        7.98&
        15.7&
        0.0&
        0.0&
       16.7 & 0.0 & 0.0&
        9.6 & 0.0 & 0.0
        \\
        \midrule
        \texttt{\tiny oursP}
        \textcolor{oursPColor}{\textbullet}&
        5.4&
        93.4&
        8.23&
        9.7&
        67.4&
        5.33&
        4.8 & 126 & 21.5&
        1.8 & 84.5 & 9.0
        \\
        \texttt{\tiny oursNoT}
        \textcolor{oursNoTColor}{\textbullet}&
        8.4&
        94.5&
        8.54&
        18.3&
        \textbf{74.7}&
        5.40&
        5.1 & 120 & 21.7 &
        1.9 & 87.0 & 9.3
        \\
        \texttt{\tiny ours}
        \textcolor{oursColor}{\textbullet}&
        12.1&
        \textbf{99.7}&
        \textbf{8.82}&
        13.3&
        72.5&
        \textbf{5.48}&
        13.6&
        \textbf{127}&
        \textbf{22.1}&
        9.4&
        \textbf{98.2}&
        \textbf{9.6}
        \\
         \bottomrule
    \end{tabular}
    \includegraphics[width = \linewidth]{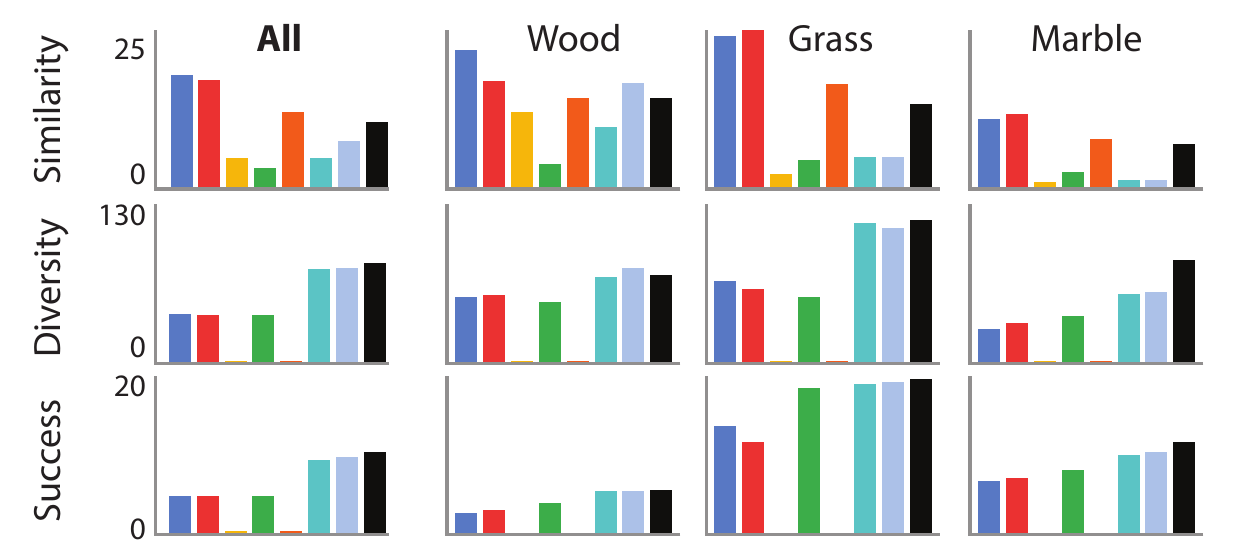}%
\end{table}

\mycfigure{Comparison}{%
Different methods and the exemplar \textbf{(columns)}, as defined in \refSec{Quantitative}, applied to different exemplars \textbf{(rows)}.
Each row shows, arranged vertically, two re-synthesises with different seeds.
Please see the text for discussion.
}

\paragraph{Fidelity}
\refFig{Quantitative} and \refTbl{Quantitative} summarize similarity, diversity and success of all methods in numbers.
\texttt{ours} method (black) comes best in diversity and success on average across all sets (first column in \refTbl{Quantitative} and top first plot in \refFig{Quantitative}).
\texttt{cnn} (yellow) and \texttt{cnnd} (green) have better similarity than any of our methods. However, no other method combines similarity with diversity as well as ours.
This is visible from the overall leading performance in the final measure, success.
This is a substantial achievement, as maximizing for only one goal is trivial: an \texttt{identity} method has zero similarity error while a \texttt{random} method has infinite diversity.

When looking at the similarity, we see that both a \texttt{cnn} and its diverse variant \texttt{cnnD} can perform similar. 
Perlin noise produces the largest error.
In particular, \texttt{perlinT} has a large error, indicating it is not sufficient to merely add a transform.
Similar, \texttt{mlp} alone cannot solve the task, as it has no access to noise and need to fit exactly, which is doable for single exemplars, but impossible for a space.
\texttt{oursNoT} has error similar to \texttt{ours}, but less diversity.

When looking at diversity, it is clear that both \texttt{cnn} and \texttt{mlp} have no diversity as they either do not have the right loss to incentivize it or have no input to generate it.
\texttt{perlin} and \texttt{perlinT} both create some level of diversity, which is not surprising as they are simple remappings of random numbers.
However, they do not manage to span the full VGG space, which only \texttt{ours} and its ablations can do.

Generating 3D textures from the exemplar in \refFig{Comparison}, we find that our diversity and  similarity are 44.5 and 1.48, which compares favorable to Perlin 3D Noise at 14.9 and 7.11.

\mysubsection{Qualitative results}{Qualitative}

Visual examples from the quantitative evaluation on a single exemplar for different methods can be seen in \refFig{Comparison}.
We see that some methods have diversity when the seed is changed (rows one vs.\ two and three vs.\ four) and some do not.
Diversity is clear for Perlin and its variant, CNNs with a diversity term and our approach.
No diversity is found for MLPs and CNNs.
We also note, that CNNs with diversity produce typically shifted copies of the same exemplar, so their diversity is over-estimated by the metric.

\myfigure{Interpolation}{Interpolation of one exemplar \textbf{(left)} into another one \textbf{(right)} in latent space (first three rows) and linear (last row).}
A meaningful latent texture code space should also allow for interpolation as seen in \refFig{Interpolation}, where we took pairs of texture codes (left and right-most exemplar) and interpolated rows in-between.
We see, that different paths produce plausible blends, with details appearing and disappearing, which is not the case for a linear blend (last row).

\mywrapfigure{InfiniteZoom}{0.25}{Zoom.}
Our method does not work on an explicit pixel grid, which allows to zoom into arbitrary fine details as show in \refFig{InfiniteZoom}, comparing favorable to cubic upsampling.
This is particularly useful in 3D, where storing a complete volume to span multiple levels of detail requires prohibitive amounts of memory while ours is output-sensitive.

\mycfigure{Stripe}{Stripes of re-synthesized textures from exemplars on the left. See the supplemental for more examples.}

\mycfigure{SolidTexturing}{%
3D texturing of different 3D shapes. Insets (right) compare ours to 2D texturing. See supplemental for 3D spin.
}

\myfigure{Reconstruction}{
Our reconstruction of \textsc{Wood}, \textsc{Grass}, \textsc{Rust}, and \textsc{Marble}.
The first row shows different input exemplars.
The second and third row show our reconstruction with different seeds.
}

\refFig{Stripe} shows a stripe re-synthesized from a single exemplar.
We note that the pattern captures the statistics, but does not repeat.

\refFig{Reconstruction} documents the ability to reproduce the entire space.
We mapped exemplars unobserved at training time to texture codes, from which we reconstruct them, in 2D.
We find that our approach reproduces the exemplars faithfully, albeit totally different on the pixel level.

Our system can construct textures and spaces of textures in 3D from 2D exemplars alone.
This is shown in \refFig{SolidTexturing}.
We first notice, that the textures have been transferred to 3D faithfully, inheriting all the benefits of procedural textures in image synthesis.
We can now take any shape, without a texture parametrization and by simply running the NN at each pixel's 3D coordinate produce a color.
We compare to a 2D approach by loading the objects in Blender and applying its state-of-the-art UV mapping approach \cite{levy2002least}.
Inevitably, a sphere will have discontinuities and poles that can not be resolved in 2D, that are no issue to our 3D approach while both take the same 2D as input.

\mysubsection{User study}{UserStudy}
Presenting $M=144$ pairs of images produced by either \texttt{perlinT}, \texttt{cnnD}, \texttt{mlp}, \texttt{oursP}, \texttt{oursNoT} and \texttt{ours} for one exemplar texture to $N=28$ subjects and asking which result ``they prefer'' in a two-alternative forced choice, we find
that
16.7\% prefer the ground truth,
4.9\% \texttt{perlin},
7.7\% \texttt{perlinT},
14.3\% \texttt{cnn},
8.8\% \texttt{cnnD},
9.4\% \texttt{mlp},
10.8\% \texttt{oursNoT},
12.9\% \texttt{oursP} and
14.5\% \texttt{ours}
(statistical significance; $p<.1$, binomial test).
Given ground truth and \texttt{cnn} are not diverse, out of all methods that synthesize infinite textures our results are preferred over all other.

\mysection{Conclusion}{Conclusion}
We have proposed a generative model of natural 3D textures.
It is trained on 2D exemplars only, and provides interpolation, synthesis and reconstruction in 3D.
The key inspiration is Perlin Noise -- now more than 30 years old -- revisited with NNs to match complex color relations in 3D according to the statistics of VGG activations in 2D.
The approach has the best combination of similarity and diversity compared to a range of published alternatives, that are less computationally efficient.

Reshaping noise to match VGG activations using MLPs can be a scalable solution to other problems in even higher dimensions, such as time, that are difficult for CNNs.

\paragraph{Acknowledgements}
This work was supported by the ERC Starting Grant SmartGeometry, a GPU donation by NVIDIA Corporation and a Google AR/VR Research Award.

\FloatBarrier
\clearpage

\bibliographystyle{ieee_fullname}
\bibliography{egbib}

\clearpage
\appendix


\mysection{Network Architecture}{NetworkArchitectures}
\mysubsection{Encoder}{Encoder}
The architecture for the encoder network remains consistent for both ours and competitor methods. Depending on training for \emph{space}, \emph{single}, \emph{w/o transform} the parameter N changes accordingly.

\begin{table}[htb]
\setlength{\tabcolsep}{1.7pt}
\center
\caption{
Network architecture for encoder.
}
\label{tbl:NetworkEncoder}
\centering
\begin{tabular}{ l  l  l  r  r }
\toprule
\textbf{Layer} & \textbf{Kernel} & \textbf{Activation} & \textbf{Shape} & \textbf{\# params} \\
\hline
Input & --- & --- & \hspace{0.25em} 3 x 128 x 128 & ---  \\
Conv & 3x3 & IN+LReLU & \hspace{0.25em} 32 x 128 x 128 & $\sim$1k \\
Conv & 4x4 & IN+LReLU &   \hspace{0.25em} 64 x \hspace{0.25em} 64 x \hspace{0.25em} 64 & $\sim$32k \\
Conv & 4x4 & IN+LReLU & 128 x \hspace{0.25em} 32 x \hspace{0.25em} 32 & $\sim$130k \\
Conv & 4x4 & IN+LReLU & 256 x \hspace{0.25em} 16 x \hspace{0.25em} 16 & $\sim$524k \\
Conv & 4x4 & IN+LReLU & 256 x \hspace{0.75em} 8 x \hspace{0.75em} 8   & $\sim$1M \\
Conv & 4x4 & IN+LReLU & 256 x \hspace{0.75em} 4 x \hspace{0.75em} 4   & $\sim$1M \\
Linear  & ---  & --- &  8   & $\sim$32k \\
Linear  & ---  & --- &  N   & $\sim$0.5k \\
\hline
\# params   & --- & ---           &                   &  $\sim$2.8M \\
\bottomrule
\end{tabular}
\end{table}

\mysubsection{Sampler}{Sampler}
The sampler architecture used for both our and the \emph{mlp} \cite{oechsle2019texturefields} method consists of following convolutional architecture with 1x1 kernels emulating Linear layers:

\begin{table}[htb]
\setlength{\tabcolsep}{1.7pt}
\center
\caption{
Network architecture for sampler.
}
\label{tbl:NetworkSampler}
\centering
\begin{tabular}{ l  l  l  r  r }
\toprule
\textbf{Layer} & \textbf{Kernel} & \textbf{Activation} & \textbf{Shape} & \textbf{\# params} \\
\hline
Input & --- & --- & \hspace{0.75em} N x 128 x 128 & ---  \\
Conv & 1x1 & ReLU &128 x 128 x 128 & $\sim$10k \\
Conv & 1x1 & ReLU &128 x 128 x 128 & $\sim$16.5k \\
Conv & 1x1 & ReLU &128 x 128 x 128 & $\sim$16.5k \\
Conv & 1x1 & ReLU &128 x 128 x 128 & $\sim$16.5k \\
Conv & 1x1 & ReLU &128 x 128 x 128 & $\sim$16.5k \\
Conv & 1x1 & ReLU & \hspace{0.75em}3 x 128 x 128 & $\sim$400 \\

\hline
\# params   & --- & ---           &                   &  $\sim$77k \\
\bottomrule
\end{tabular}
\end{table}

\mysubsection{CNN}{CNN}
For \emph{cnn} and \emph{cnnD} competitors we use a similar architecture to the proposed method of \cite{ulyanov2017improved}:

\begin{table}[htb]
\setlength{\tabcolsep}{1.7pt}
\center
\caption{
Network architecture for convolutional methods.
}
\label{tbl:NetworkCNN}
\centering
\begin{tabular}{ l  l  l  r  r }
\toprule
\textbf{Layer} & \textbf{Kernel} & \textbf{Activation} & \textbf{Shape} & \textbf{\# params} \\
\hline
Input & --- & --- &  (32) + 256 & ---  \\
Linear & --- & --- & (32) + 256 & $\sim$80k \\
Linear & --- & --- & 256 & $\sim$70k \\
Reshape & --- & --- &  16 x \hspace{1em}4 x \hspace{1em}4 & --- \\
ConvT & 4x4 & ReLU & 128 x \hspace{1em}8 x \hspace{1em}8 & $\sim$32k \\
ConvT & 4x4 & ReLU & 128 x \hspace{0.5em}16 x \hspace{0.5em}16 & $\sim$260k \\
ConvT & 4x4 & ReLU & 128 x \hspace{0.5em}32 x \hspace{0.5em}32 & $\sim$260k \\
Upsample & --- & --- &  128 x \hspace{0.5em}64 x \hspace{0.5em}64 & ---\\
Conv & 3x3 & ReLU & 64 x \hspace{0.5em}64 x \hspace{0.5em}64 & $\sim$70k \\
Upsample & --- & --- &  64 x 128 x 128 & ---\\
Conv & 3x3 & ReLU & 3 x 128 x 128 & $\sim$2k \\

\hline
\# params   & --- & ---           &                   &  $\sim$790k \\
\bottomrule
\end{tabular}
\end{table}

\newpage

\mysection{Results}{Results}
Additional results of stripe images and interpolations are displayed below.

A webpage containing more results for all four classes (\textsc{Wood}, \textsc{Marble}, \textsc{Grass} and \textsc{Rust}) including competitors can be accessed online: \url{https://geometry.cs.ucl.ac.uk/projects/2020/neuraltexture}. Additionally, videos of rotating shapes textured by our method are provided.
Our code is available at: \url{https://github.com/henzler/neuraltexture}

\mycfigure{WoodSpace}{Results derived from the encoded \textsc{Wood} space.}
\mycfigure{MarbleSpace}{Results derived from the encoded \textsc{Marble} space.}
\mycfigure{GrassSpace}{Results derived from the encoded \textsc{Grass} space.}
\mycfigure{RustSpace}{Results derived from the encoded \textsc{Rust} space.}

\mycfigure{WoodInterpolated}{Latent space interpolation from one ground truth wood exemplar (left) into secondary ground truth exemplar (right). Each row corresponds to independent interpolations.}

\mycfigure{GrassInterpolated}{Latent space interpolation from one ground truth grass exemplar (left) into secondary ground truth exemplar (right). Each row corresponds to independent interpolations.}
\mycfigure{MarbleInterpolated}{Latent space interpolation from one ground truth marble exemplar (left) into secondary ground truth exemplar (right). Each row corresponds to independent interpolations.}
\mycfigure{RustInterpolated}{Latent space interpolation from one ground truth rust exemplar (left) into secondary ground truth exemplar (right). Each row corresponds to independent interpolations.}

\end{document}